\documentclass{article}

\usepackage[final]{neurips_2019}

\usepackage[utf8]{inputenc} 
\usepackage[T1]{fontenc}    
\usepackage{hyperref}       
\usepackage{url}            
\usepackage{booktabs}       
\usepackage{amsfonts}       
\usepackage{nicefrac}       
\usepackage{microtype}      

\usepackage{amssymb}
\RequirePackage{amsmath}
\usepackage{algorithm}
\usepackage{enumitem}
\usepackage{hyperref}
\usepackage{subfig}
\usepackage[noend]{algpseudocode}
\makeatletter
\def\BState{\State\hskip-\ALG@thistlm}
\makeatother
\usepackage{graphicx}
\usepackage{wrapfig}
\usepackage{epstopdf}
\usepackage{todonotes}
\def\HS{\hspace{\fontdimen2\font}}

\definecolor{citecolor}{RGB}{0,77,128}
\hypersetup{colorlinks,
linkcolor=citecolor,
citecolor=citecolor
}

\title{MERL: Multi-Head Reinforcement Learning}

\author{%
  Yannis Flet-Berliac {\normalfont and} Philippe Preux \\
  \\ SequeL Inria, University of Lille\\ CRIStAL, CNRS
}

\begin{document}

\maketitle

\begin{abstract}
  A common challenge in reinforcement learning is how to convert the agent's interactions with an environment into fast and robust learning. For instance, earlier work makes use of domain knowledge to improve existing reinforcement learning algorithms in complex tasks. While promising, previously acquired knowledge is often costly and challenging to scale up. Instead, we decide to consider problem knowledge with signals from quantities relevant to solve any task, e.g., self-performance assessment and accurate expectations. $\mathcal{V}^{ex}$ is such a quantity. It is the fraction of variance explained by the value function $V$ and measures the discrepancy between $V$ and the returns. Taking advantage of $\mathcal{V}^{ex}$, we propose MERL, a general framework for structuring reinforcement learning by injecting problem knowledge into policy gradient updates. As a result, the agent is not only optimized for a reward but learns using problem-focused quantities provided by MERL, applicable out-of-the-box to any task. In this paper: (a) We introduce and define MERL, the multi-head reinforcement learning framework we use throughout this work. (b) We conduct experiments across a variety of standard benchmark environments, including 9 continuous control tasks, where results show improved performance. (c) We demonstrate that MERL also improves transfer learning on a set of challenging pixel-based tasks. (d) We ponder how MERL tackles the problem of reward sparsity and better conditions the feature space of reinforcement learning agents.
\end{abstract}

\section{Introduction}
The problem of learning how to act optimally in an unknown dynamic environment has been a source of many research efforts for decades~\citep{nguyen1990truck,werbos1989neural,schmidhuber1991learning} and is still at the forefront of recent work in deep Reinforcement Learning (RL)~\citep{burda2018exploration,ha2018recurrent,silver_mastering_2016,espeholt2018impala}. Nevertheless, current algorithms tend to be fragile and opaque~\citep{iyer2018transparency}: they require a large amount of training data collected from an agent interacting with a simulated environment where the reward signal is often critically sparse. Collecting signals that will make the agent more efficient is, therefore, at the core of the algorithms designers' concerns.

Previous work in RL uses prior knowledge~\citep{lin1992self,clouse1992teaching,ribeiro1998embedding,moreno2004using} to reduce sample inefficiency. While promising and unquestionably necessary, the integration of such priors into current methods is likely costly to implement, it may cause undesired constraints and can hinder scaling up. Therefore, we propose a framework to directly integrate \textit{non-limiting constraints} in current RL algorithms while being applicable to any task. In addition to an increased efficiency, the agent should learn from all interactions, not just the rewards. Indeed, if the probability of receiving a reward by chance is arbitrarily low, then the time required to learn from it will be arbitrarily long~\citep{whitehead1991complexity}. This barrier to learning will prevent agents from significantly reducing learning time. One way to overcome this barrier is to learn \textit{complementary and task-agnostic signals} of self-performance assessment and accurate expectations from different sources~\citep{schmidhuber1991curious,oudeyer2007intrinsic}, whatever the task to master.
\begin{wrapfigure}[29]{r}{0.5\textwidth}
\centering
\includegraphics[width=\linewidth]{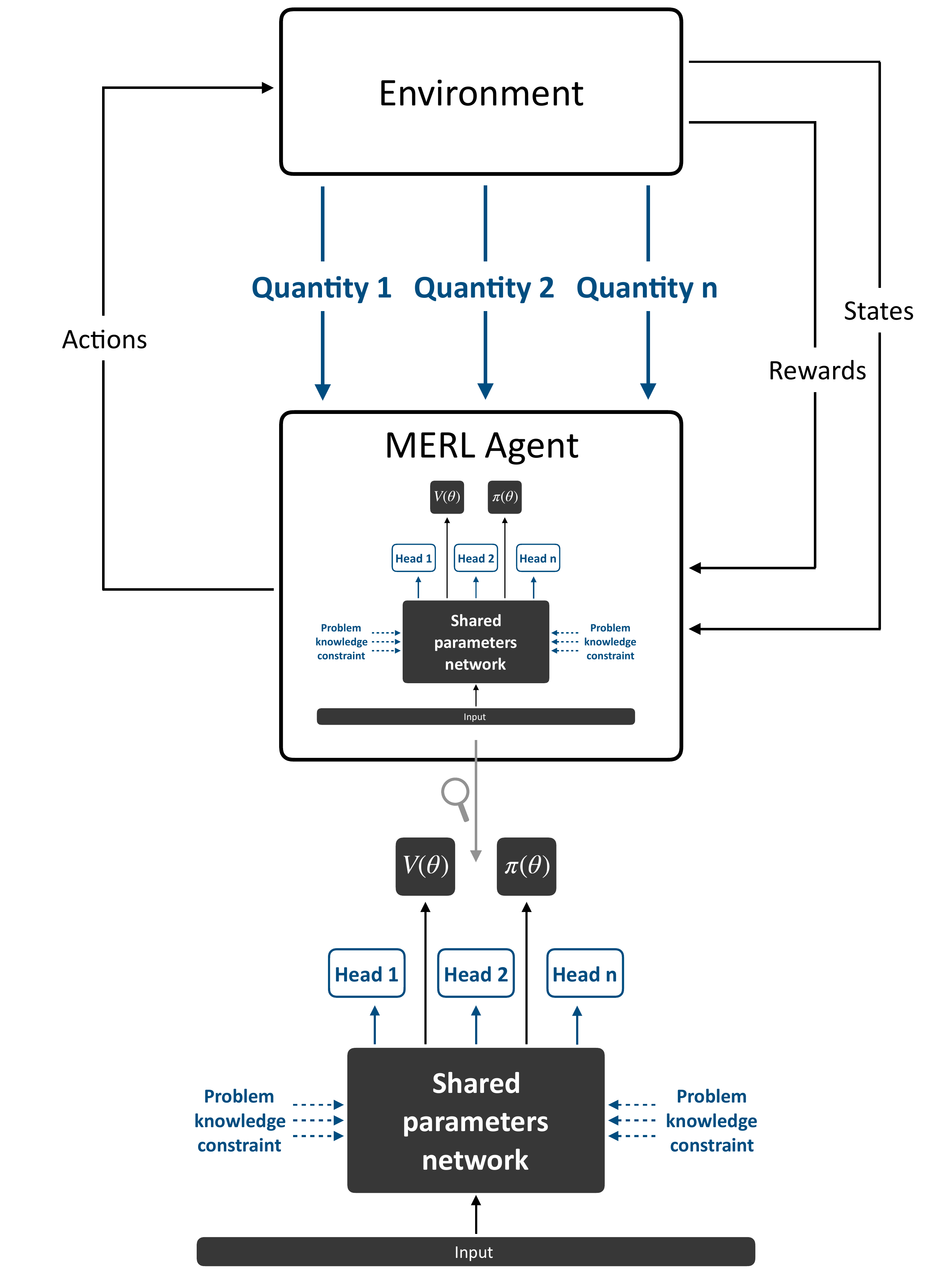}
\caption{High-level overview of the Multi-hEad Reinforcement Learning (MERL) framework.}
\label{fig:overview}
\end{wrapfigure}

From the above considerations and building on existing auxiliary task methods, we design a framework that integrates problem knowledge quantities into the learning process. In addition to providing a method technically applicable to any policy gradient method or environment, the central idea of MERL is to incorporate a measure of the discrepancy between the estimated state value and the observed returns as an auxiliary task. This discrepancy is formalized with the notion of the fraction of variance explained $\mathcal{V}^{ex}$~\citep{10.2307/2683704}. One intuition the reader can have is that MERL transforms a reward-focused task into a task regularized with dense problem knowledge signals.

Fig.\@~\ref{fig:overview} provides a preliminary understanding of MERL assets: an enriched actor-critic architecture with a lightly modified learning algorithm places the agent amidst task-agnostic auxiliary quantities directly sampled from the environment. In the sequel of this paper, we use two problem knowledge quantities to demonstrate the performance of MERL: $\mathcal{V}^{ex}$, a compelling measure of self-performance, and future states prediction, commonly used in auxiliary task methods. The reader is further encouraged to introduce many other relevant signals. We demonstrate that while being able to predict the quantities from the different MERL heads correctly, the agent outperforms the on-policy baseline that does not use the MERL framework on various continuous control tasks. We also show that, interestingly, our framework allows to better transfer the learning, from one task to another, on several \textit{Atari 2600} games.

\section{Preliminaries}
We consider a Markov Decision Process (MDP) with states $s \in \mathcal{S}$, actions $a \in \mathcal{A}$, transition distribution $s_{t+1} \sim P(s_{t},a_{t})$ and reward function $r(s,a)$. Let $\pi=\{\pi(a | s), s \in \mathcal{S}, a \in \mathcal{A}\}$ denote a stochastic policy and let the objective function be the traditional expected discounted reward:
\begin{equation}
J(\pi) \triangleq \underset{\tau \sim \pi}{\mathbb{E}}\left[\sum_{t=0}^{\infty} \gamma^{t} r\left(s_{t}, a_{t}\right)\right],
\end{equation}
where $\gamma \in [0,1)$ is a discount factor~\citep{puterman1994markov} and $\tau=\left(s_{0}, a_{0}, s_{1}, \dots\right)$ is a trajectory sampled from the environment. Policy gradient methods aim at modelling and optimizing the policy directly~\citep{williams1992simple}. The policy $\pi$ is generally implemented with a function parameterized by $\theta$. In the sequel, we will use $\theta$ to denote the parameters as well as the policy. In deep RL, the policy is represented in a neural network called the policy network and is assumed to be continuously differentiable with respect to its parameters $\theta$.

\subsection{Fraction of Variance Explained: $\mathcal{V}^{ex}$}
The fraction of variance that the value function explains about the returns corresponds to the proportion of the variance in the dependent variable $V$ that is predictable from the independent variable $s_{t}$. We define $\mathcal{V}^{ex}_{\tau}$ as the \textit{fraction of variance explained} for a trajectory $\tau$:
\begin{equation}
    \mathcal{V}^{ex}_{\tau} \triangleq 1 - \frac{\sum_{t\in\tau}\left(\hat{R}_{t}-V(s_{t})\right)^{2}}{\sum_{t\in\tau}\left(\hat{R}_{t}-\overline{R}\right)^{2}},
    \label{eq:vex}
\end{equation}
where $\hat{R}_{t}$ and $V(s_{t})$ are respectively the return and the expected return from state $s_{t} \in \tau$, and $\overline{R}$ is the mean of all returns in trajectory $\tau$. In statistics, this quantity is also known as the coefficient of determination $\mathit{R}^{2}$ and it should be noted that this criterion may be negative for non-linear models~\citep{10.2307/2683704}, indicating a severe lack of fit of the corresponding function:
\begin{itemize}
    \item $\mathcal{V}^{ex}_{\tau} = 1$: $V$ perfectly explains the returns - $V$ and the returns are \textit{correlated};
    \item $\mathcal{V}^{ex}_{\tau} = 0$ corresponds to a simple average prediction - $V$ and the returns are \textit{not correlated};
    \item $\mathcal{V}^{ex}_{\tau} < 0$: $V$ provides a worse fit to the outcomes than the mean of the returns.
\end{itemize}

One can have the intuition that $\mathcal{V}^{ex}_{\tau}$ close to 1 implies that the trajectory ${\tau}$ provides valuable signals because they correspond to transitions sampled from an exercised behavior. On the other hand, $\mathcal{V}^{ex}_{\tau}$ close to 0 indicates that the value function is not correlated with the returns, therefore, the corresponding samples are not expected to provide as valuable information as before. Finally, $\mathcal{V}^{ex}_{\tau} < 0$ corresponds to a high mean-squared error for the value function, which means for the related trajectory that the agent still has to learn to perform better. In~\citet{flet2019samples}, policy gradient methods are improved by using $\mathcal{V}^{ex}$ as a criterion to dropout transitions before each policy update. We will show that $\mathcal{V}^{ex}$ is also a relevant indicator for assessing self-performance in the context of MERL agents.

\subsection{Policy Gradient Method: PPO with Clipped Surrogate Objective}
In this paper, we consider on-policy learning primarily for its unbiasedness and stability compared to off-policy methods~\citep{nachum2017bridging}. On-policy is also empirically known as being less sample efficient than off-policy learning hence this issue emerged as an interesting research topic. However, our method can be applied to off-policy methods as well, and we leave this investigation open for future work.

PPO~\citep{schulman2017proximal} is among the most commonly used and state-of-the-art on-policy policy gradient methods. Indeed, PPO has been tested on a set of benchmark tasks and has proven to produce impressive results in many cases despite a relatively simple implementation. For instance, instead of imposing a hard constraint like TRPO~\citep{schulman2015trust}, PPO formalizes the constraint as a penalty in the objective function. In PPO, at each iteration, the new policy $\theta_{new}$ is obtained from the old policy $\theta_{old}$:

\begin{equation}
  \theta_{new} \gets \underset{\theta}{\mathrm{argmax}} \underset{s_{t}, a_{t} \sim \pi_{\theta_{old}}}{\mathbb{E}}\left[L^{\mathrm{PPO}}\left(s_{t}, a_{t}, \theta_{old}, \theta\right)\right].
\end{equation}
  We use the clipped version of PPO whose objective function is:
\begin{equation}
    L^{\mathrm{PPO}}(s_{t},a_{t},\theta_{old},\theta) = \min\left(
    \frac{\pi_{\theta}(a_{t}|s_{t})}{\pi_{\theta_{old}}(a_{t}|s_{t})}  A^{\pi_{\theta_{old}}}(s_{t},a_{t}),\;g(\epsilon, A^{\pi_{\theta_{old}}}(s_{t},a_{t}))\right),
    \label{eq:min}
\end{equation}
    where
\begin{equation}
    g(\epsilon, A) = \left\{
    \begin{array}{ll}
    (1 + \epsilon) A, A \geq 0 \\
    (1 - \epsilon) A, A < 0.
    \end{array}
    \right.
\end{equation}
$A$ is the advantage function, $A (s, a) \triangleq Q (s, a) - V (s)$. The expected advantage function $A^{\pi_{\theta_{old}}}$ is estimated by an old policy and then re-calibrated using the probability ratio between the new and the old policy. In Eq.~\ref{eq:min}, this ratio is constrained to stay within a small interval around 1, making the training updates more stable.

\subsection{Related Work}
Auxiliary tasks have been adopted to facilitate representation learning for decades~\citep{suddarth1990rule,klyubin2005empowerment}, along with intrinsic motivation~\citep{schmidhuber2010formal,pathak2017curiosity} and artificial curiosity~\citep{schmidhuber1991curious,oudeyer2007intrinsic}. The use of auxiliary tasks to allow the agents to maximize other pseudo-reward functions simultaneously has been studied in a number of previous work~\citep{shelhamer2016loss,dosovitskiy2016learning,burda2018large,du2018adapting,riedmiller2018learning,kartal2019terminal}, including incorporating unsupervised control tasks and reward predictions in the UNREAL framework~\citep{jaderberg2016reinforcement}, applying auxiliary tasks to navigation problems~\citep{mirowski2016learning}, or for utilizing representation learning~\citep{lesort2018state} in the context of model-based RL. Lastly, in imitation learning of sequences provided by experts,~\citet{li2015recurrent} introduces a supervised loss for fitting a recurrent model on the hidden representations to predict the next observed state.

Our method incorporates two key contributions: a multi-head layer with auxiliary task signals both environment-agnostic and technically applicable to any policy gradient method, and the use of $\mathcal{V}^{ex}_{\tau}$ as an auxiliary task to measure the discrepancy between the value function and the returns in order to allow for better self-performance assessment and eventually more efficient learning. In addition, MERL differs from previous approaches in that its framework simultaneously addresses the advantages mentioned hereafter: (a) neither the introduction of new neural networks (e.g., for memory) nor the introduction of a replay buffer or an off-policy setting is needed, (b) all relevant quantities are compatible with any task and is not limited to pixel-based environments, (c) no additional iterations are required, and no modification to the reward function of the policy gradient algorithms it is applied to is necessitated. The above reasons make MERL generally applicable and technically suitable out-of-the-box to most policy gradient algorithms with a negligible computational cost overhead.

From a different perspective,~\citet{garcia2015comprehensive} gives a detailed overview of previous work that has changed the optimality criterion as a safety factor. But most methods use a hard constraint rather than a penalty; one reason is that it is difficult to choose a single coefficient for this penalty that works well for different problems. We are successfully addressing this problem with MERL. In~\citet{lipton2016combating}, catastrophic actions are avoided by training an intrinsic fear model to predict whether a disaster will occur and using it to shape rewards. Compared to both methods, MERL is more scalable and lightweight while it successfully incorporates quantities of self-performance assessments (e.g., variance explained of the value function) and accurate expectations (e.g., next state prediction) leading to an improved performance.

\section{Multi-Head Framework for Reinforcement Learning using $\mathcal{V}^{ex}$}
Our multi-head architecture and its associated learning algorithm are directly applicable to most state-of-the-art policy gradient methods. Let $h$ be the index of each MERL head: $\mathrm{MERL}^{h}$. We propose two of the quantities predicted by these heads and show how to incorporate them into PPO.

\subsection{Policy and Value Function Representation}
In deep RL, the policy is generally represented in a neural network called the policy network, with parameters $\theta$, and the value function is parameterized by the value network, with parameters $\phi$. Each MERL head $\mathrm{MERL}^{h}$ takes as input the last embedding layer from the value network and is constituted of only one layer of fully-connected neurons, with parameters $\phi^{h}$. The output size of each head corresponds to the size of the predicted MERL quantity. Below, we elaborate on two.

\subsection{$\mathcal{V}^{ex}$ Estimation}
In order to have an estimate of the fraction of variance explained, we write $\mathrm{MERL}^{\mathrm{VE}}$ as the corresponding MERL head with parameters $\phi^{\mathrm{VE}}$. Its objective function is defined by:
\begin{equation}
    L^{\mathrm{MERL}^{\mathrm{VE}}}(\tau,\phi,\phi^{\mathrm{VE}}) = \|\mathrm{MERL}^{\mathrm{VE}}(\tau)-\mathcal{V}^{ex}_{\tau}\|_{2}^{2}.
\end{equation}

\subsection{Future States Estimation}
Auxiliary task methods based on next state prediction are, to the best of our knowledge, the most commonly used in the RL literature. We include such auxiliary task into MERL, in order to assimilate our contribution to the previous work and to provide a enriched evaluation of the proposed framework. At each timestep, one of the agent's MERL heads predicts a future state $s^{\prime}$ from $s$. While a typical MERL quantity can be fit by regression on mean-squared error, we observed that predictions of future states are better fitted with a cosine-distance error. We denote $\mathrm{MERL}^{\mathrm{FS}}$ the corresponding head, with parameters $\phi^{\mathrm{FS}}$, and $S$ the observation space size (size of vector $s$). We define its objective function as:
\begin{equation}
    L^{\mathrm{MERL}^{\mathrm{FS}}}(s,\phi,\phi^{\mathrm{FS}}) = 1 - \frac{\sum_{i=1}^{S} \mathrm{MERL}^{\mathrm{FS}}_{i}(s) \cdot s_{i}^{\prime}}{\sqrt{\sum_{i=1}^{S} (\mathrm{MERL}^{\mathrm{FS}}_{i}(s))^{2}} \sqrt{\sum_{i=1}^{S} (s_{i}^{\prime})^{2}}}.
\end{equation}

\subsection{Problem-Constrained Policy Update}
Once a set of MERL heads $\mathrm{MERL}^{h}$ and their associated objective functions $L^{\mathrm{MERL}^{h}}$ have been defined, we modify the gradient update step of the policy gradient algorithms. The objective function incorporates all $L^{\mathrm{MERL}^{h}}$. Of course, each MERL objective is associated with its coefficient $c_{h}$. It is worthy to note that we used the exact same MERL coefficients for all our experiments, which demonstrate the framework's ease of applicability. Algorithm~\ref{alg:ppo} illustrates how the learning is achieved. In Eq.~\ref{eq:ppo_value}, only the (boxed) MERL objectives parameterized by $\phi$ are added to the value update and modify the learning algorithm.
\begin{algorithm}
\caption{PPO+MERL update.}\label{alg:ppo}
\begin{algorithmic}
\State \textbf{Initialise} policy parameters $\theta_{0}$
\State \textbf{Initialise} value function and $\mathrm{MERL}^{h}$ functions parameters $\phi_{0}$
\State
\For {k = 0,1,2,...}
\State \textbf{Collect} \textit{set of trajectories} $\mathcal{D}_{k}=\left\{\tau_{i}\right\}$ with horizon $T$ by running policy $\pi_{\theta_{k}}$ in the environment
\State \textbf{Compute} \textit{$\mathrm{MERL}^{h}$ estimates} at timestep $t$ from sampling the environment
\State \textbf{Compute} \textit{advantage estimates} $A_{t}$ at timestep $t$ based on the current value function $V_{\phi_{k}}$
\State \textbf{Compute} \textit{future rewards} $\hat{R}_{t}$ from timestep $t$
\State
\State \textbf{Gradient Update}
\begin{align}
\theta_{k+1} =\HS\HS\HS &\underset{\theta}{\mathrm{argmax}} \sum_{\tau \in \mathcal{D}_{k}}\sum_{t=0}^{T} \min \left(\frac{\pi_{\theta}\left(a_{t} | s_{t}\right)}{\pi_{\theta_{k}}\left(a_{t} | s_{t}\right)} A^{\pi_{\theta_{k}}}\left(s_{t}, a_{t}\right), g\left(\epsilon, A^{\pi_{\theta_{k}}}\left(s_{t}, a_{t}\right)\right)\right) \label{eq:ppo_policy}
 \\
\phi_{k+1} =\HS\HS\HS &\underset{\phi}{\mathrm{argmin}} \sum_{\tau \in \mathcal{D}_{k}}\sum_{t=0}^{T}\left(V_{\phi_{k}}\left(s_{t}\right)-\hat{R}_{t}\right)^{2} + \boxed{ \sum_{h=0}^{H}c_{h}L^{\mathrm{MERL}^{h}}} \label{eq:ppo_value}
\end{align}

\EndFor
\end{algorithmic}
\end{algorithm}

\section{MERL applied to Continuous Control Tasks and the Atari domain}
\subsection{Methodology}
We evaluate MERL in multiple high-dimensional environments, ranging from \textit{MuJoCo}~\citep{todorov2012mujoco} to the \textit{Atari 2600} games~\citep{bellemare2013arcade}. The experiments in \textit{MuJoCo} allow us to evaluate the performance of MERL on a large number of different continuous control problems. It is worthy to note that the universal characteristics of the auxiliary quantities we design ensure that MERL is directly applicable to any task. Other popular auxiliary task methods~\citep{jaderberg2016reinforcement,mirowski2016learning,burda2018large} are not out-of-the-box applicable to continuous control tasks like \textit{MuJoCo}. Thus, we naturally compare the performance of our method with PPO~\citep{schulman2017proximal} where MERL heads are not used. Later, we also experiment with MERL on the \textit{Atari 2600} games to study the transfer learning abilities of our method on a set of diverse tasks.

\begin{description}[leftmargin=0pt]
  \item[Implementation.] For the continuous control \textit{MuJoCo} tasks, the agents have learned using separated policy and value networks. In this case, we build upon the value network to incorporate our framework's heads. On the contrary, when playing \textit{Atari 2600} games from pixels, the agents were given a CNN network~\citep{krizhevsky2012imagenet} shared between the policy and the value function. In that case, $\mathrm{MERL}^{h}$ are naturally attached to the last embedding layer of the shared network. In both configurations, the outputs of $\mathrm{MERL}^{h}$ heads are the same size as the quantity they predict: for instance, $\mathrm{MERL}^{\mathrm{VE}}$ is a scalar whereas $\mathrm{MERL}^{\mathrm{FS}}$ is a state.
\end{description}

\begin{description}[leftmargin=0pt]
  \item[Hyper-parameters Setting.] We used the same hyper-parameters as in the main text of the corresponding paper. We made this choice within a clear and objective protocol of demonstrating the benefits of using MERL. Hence, its reported performance is not necessarily the best that can be obtained, but it still exceeds the baseline. Using MERL adds as many hyper-parameters as there are heads in the multi-head layer and it is worth noting that MERL hyper-parameters are the same for all tasks. We report all hyper-parameters in Tables~\ref{tab:hyper1} and~\ref{tab:hyper2}.
\end{description}

\begin{table}[!ht]
\caption{Hyper-parameters used in PPO+MERL}
\label{tab:hyper1}
\centering
\setlength{\tabcolsep}{8pt}
\begin{tabular}{l | l}

Hyper-parameter                        & Value             \\ \hline
Horizon ($T$)               & 2048 (MuJoCo), 128 (Atari)              \\
Adam stepsize                         & $3 \cdot 10^{-4}$ (MuJoCo), $2.5 \cdot 10^{-4}$ (Atari) \\
Nb. epochs                            & 10 (MuJoCo), 3 (Atari)                \\
Minibatch size                        & 64 (MuJoCo), 32 (Atari)             \\
Number of actors                    & 1 (MuJoCo), 4 (Atari) \\
Discount ($\gamma$)                   & 0.99              \\
GAE parameter ($\lambda$)             & 0.95              \\
Clipping parameter ($\epsilon$)       & 0.2 (MuJoCo), 0.1 (Atari)               \\
Value function coef                     & 0.5               \\
\bottomrule
\end{tabular}
\end{table}

\begin{table}[!ht]
\caption{MERL hyper-parameters}
\label{tab:hyper2}
\centering
\setlength{\tabcolsep}{8pt}
\begin{tabular}{l | l}

Hyper-parameter                        & Value             \\ \hline
$\mathrm{MERL}^{\mathrm{VE}}$ coef $c_{\mathrm{VE}}$     & 0.5               \\
$\mathrm{MERL}^{\mathrm{FS}}$ coef $c_{\mathrm{FS}}$ & 0.01               \\
\bottomrule
\end{tabular}
\end{table}

\begin{description}[leftmargin=0pt]
  \item[Performance Measures.] We examine the performance across a large number of trials (with different seeds for each task). Standard deviation of returns, and average return are generally considered to be the most stable measures used to compare the performance of the algorithms being studied~\citep{islam2017reproducibility}. Thereby, in the rest of this work, we use those metrics to establish the performance of our framework quantitatively.
\end{description}

\subsection{Single-Task Learning: Continuous Control}
\begin{figure}[h]
    \centering
    \subfloat{{\includegraphics[width=.33\linewidth]{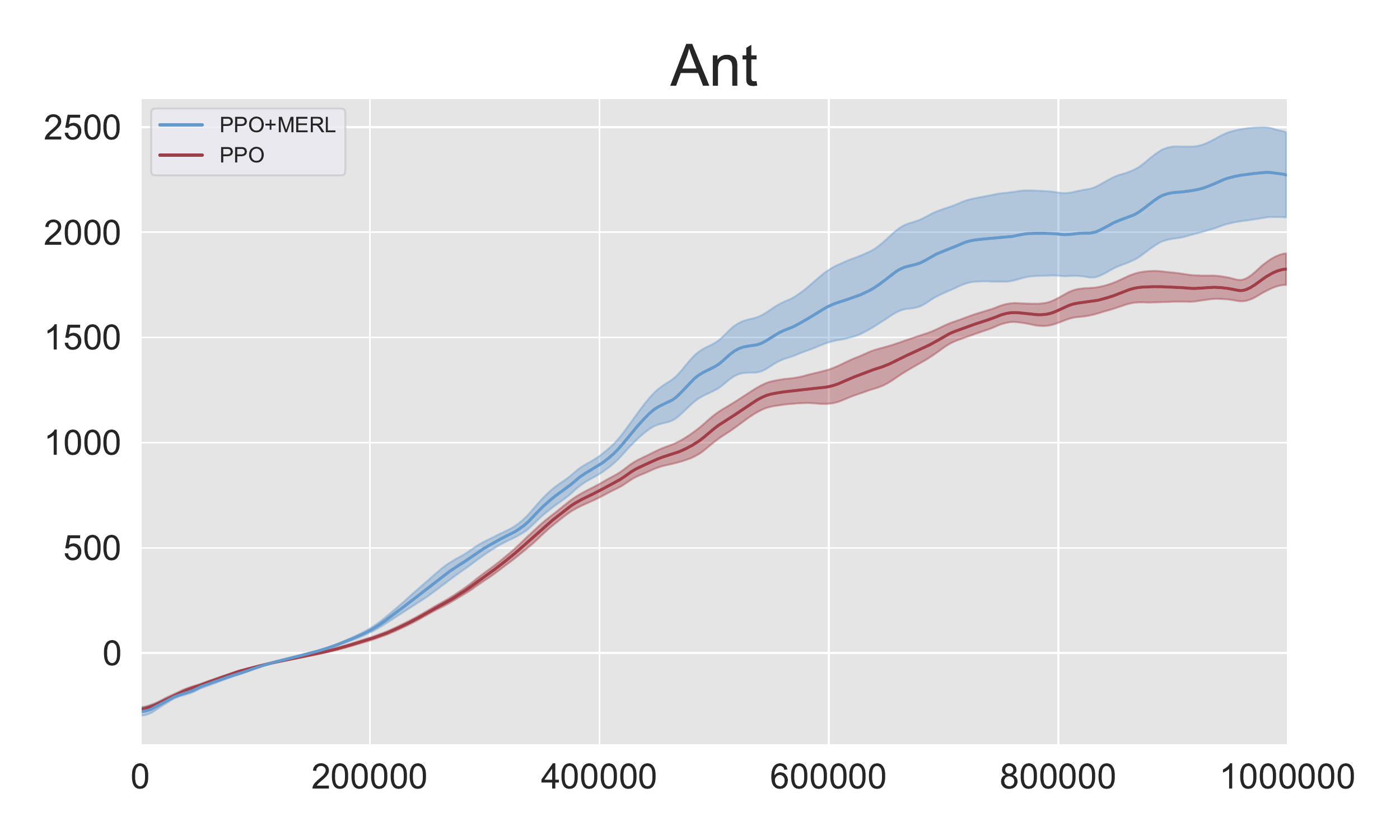}}{\includegraphics[width=.33\linewidth]{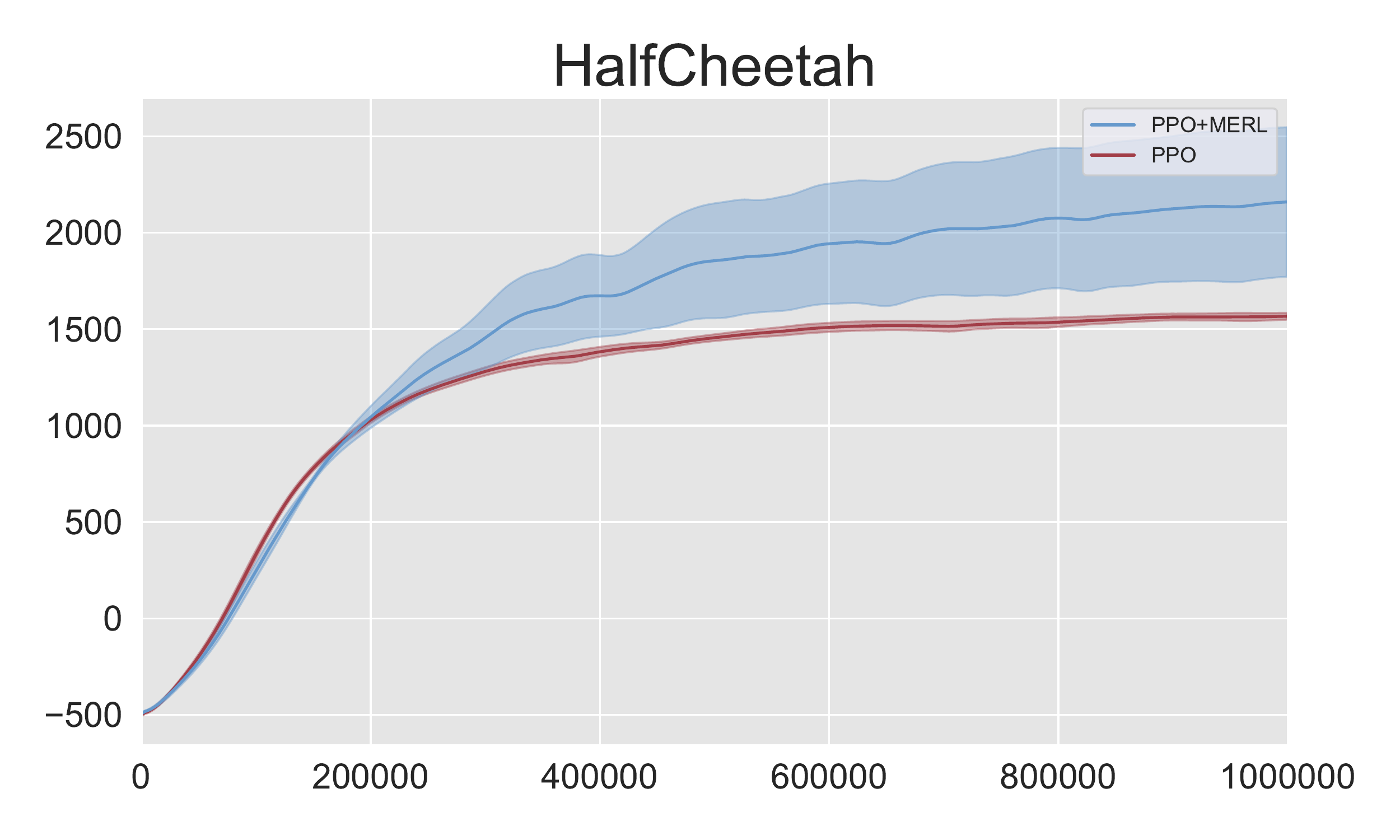}}{\includegraphics[width=.33\linewidth]{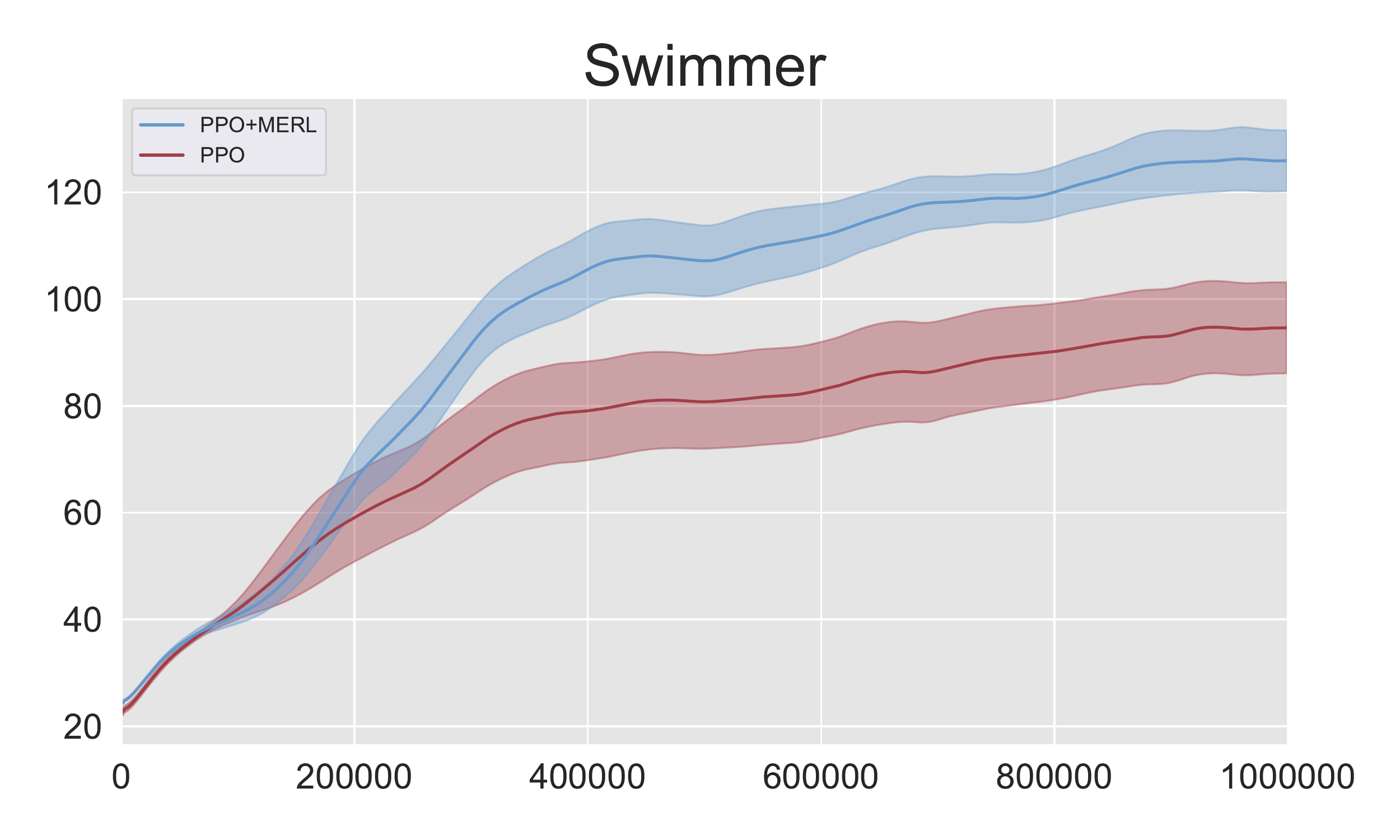}}}
    \caption{Experiments on 3 MuJoCo environments ($10^6$ timesteps, 7 seeds) with PPO+MERL. Red is the baseline, blue is with our method. The line is the average performance, while the shaded area represents its standard deviation.}
    \label{fig:mujoco_ppo}
\end{figure}

We apply MERL to PPO in several continuous control tasks, where using auxiliary tasks has not been explored in detail in the literature. Specifically, we use 9 \textit{MuJoCo} environments. Due to space constraints, only 3 graphs from varied tasks are shown in Fig.\@~\ref{fig:mujoco_ppo}. The complete set of 9 tasks is reported in Table~\ref{tab:res}.

\begin{table}[h]
  \centering
  \caption{Average total reward of the last 100 episodes over 7 runs on the 9 MuJoCo environments. \textbf{Boldface} $mean\pm std$ indicate statistically better performance.}
  \label{tab:res}
  \setlength{\tabcolsep}{8pt}
  \begin{tabular}{l | l | l}
Task                     & PPO          & Ours                      \\ \hline
Ant                      & $1728\pm64$  & $\mathbf{2157\pm212}$    \\
HalfCheetah              & $1557\pm21$  & $\mathbf{2117\pm370}$    \\
Hopper                   & $2263\pm125$  & $2105\pm200$     \\
Humanoid                 & $577\pm10$  & $\mathbf{603\pm8}$    \\
InvertedDoublePendulum   & $5965\pm108$  & $\mathbf{6604\pm130}$        \\
InvertedPendulum         & $474\pm14$  & $\mathbf{497\pm12}$        \\
Reacher                  & $-7.84\pm0.7$  & $-7.78\pm0.8$    \\
Swimmer                  & $93.2\pm8.7$  & $\mathbf{124.6\pm5.6}$ \\
Walker2d                 & $2309\pm332$  & $2347\pm353$ \\
\end{tabular}
\end{table}

The results demonstrate that using MERL leads to better performance on a variety of continuous control tasks. Moreover, learning seems to be faster for some tasks, suggesting that MERL takes advantage of its heads to learn relevant quantities from the beginning of learning, when the reward signals may be sparse. Interestingly, by looking at the performance across all 9 tasks, we observed better results by predicting only the next state and not the subsequent ones.

\subsection{Transfer Learning: Atari Domain}
Because of training time constraints, we consider a transfer learning setting where, after the first $10^6$ training steps, the agent switches to a new task for another $10^6$ steps. The agent is not aware of the task switch.~\textit{Atari 2600} has been a challenging testbed for many years due to its high-dimensional video input (210 x 160) and the discrepancy of tasks between games. To investigate the advantages of MERL in transfer learning, we choose a set of 6 Atari games with an action space of 9, which is the average size of the action space in the Atari domain. This experimental choice is beneficial in that the 6 games provide a diverse range of game-play while sticking to the same size of action space.

\begin{figure}[h]
    \centering
    \subfloat{{\includegraphics[width=.33\linewidth]{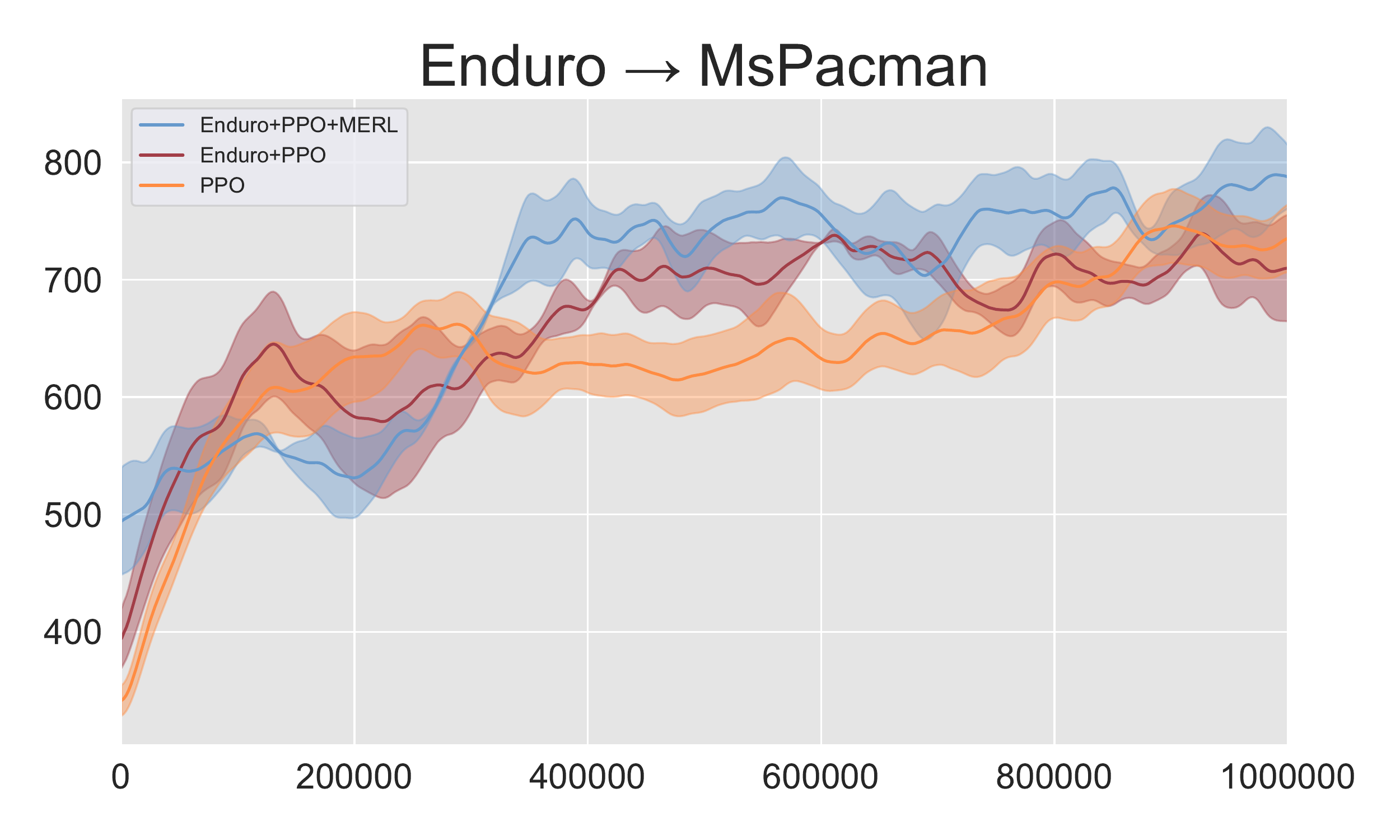}}{\includegraphics[width=.33\linewidth]{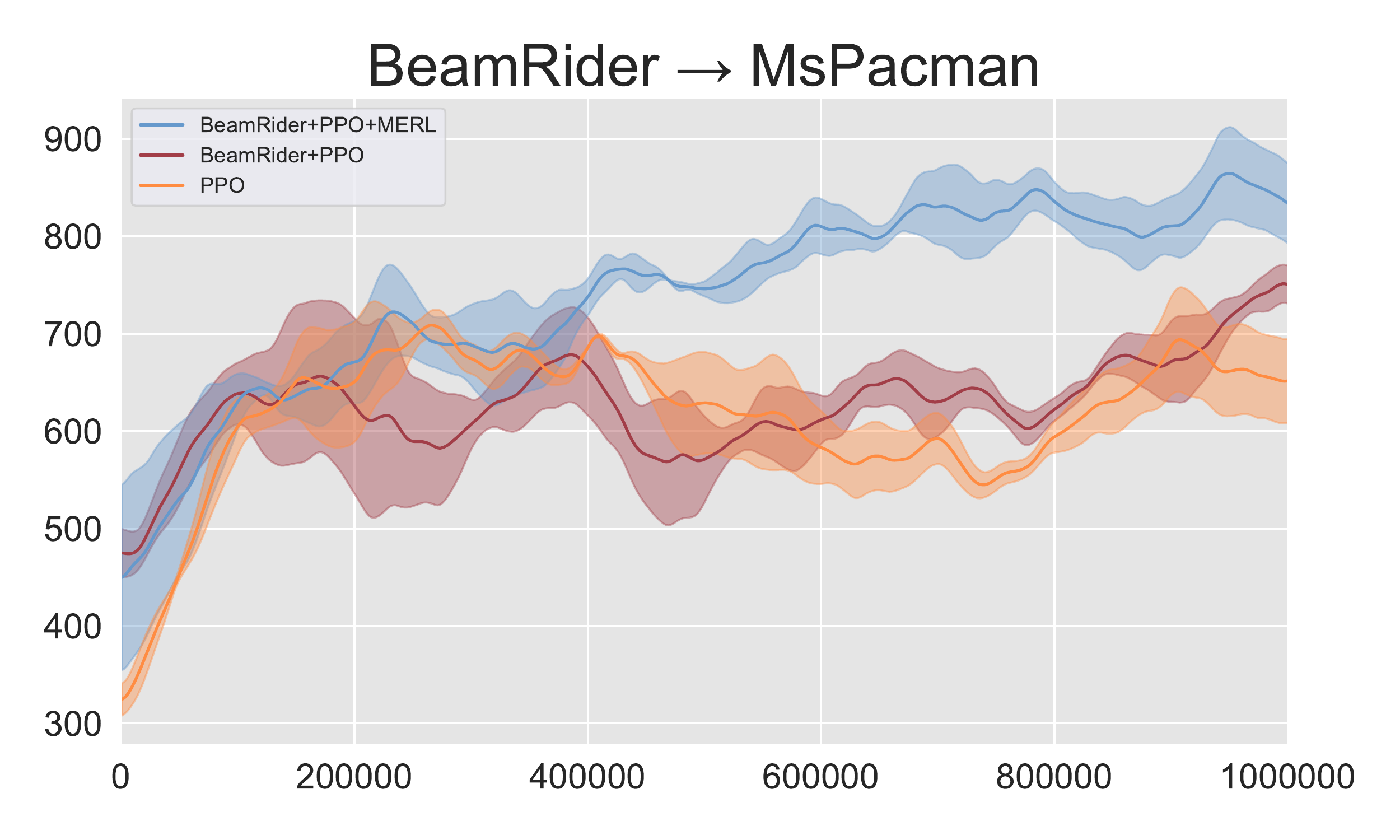}}{\includegraphics[width=.33\linewidth]{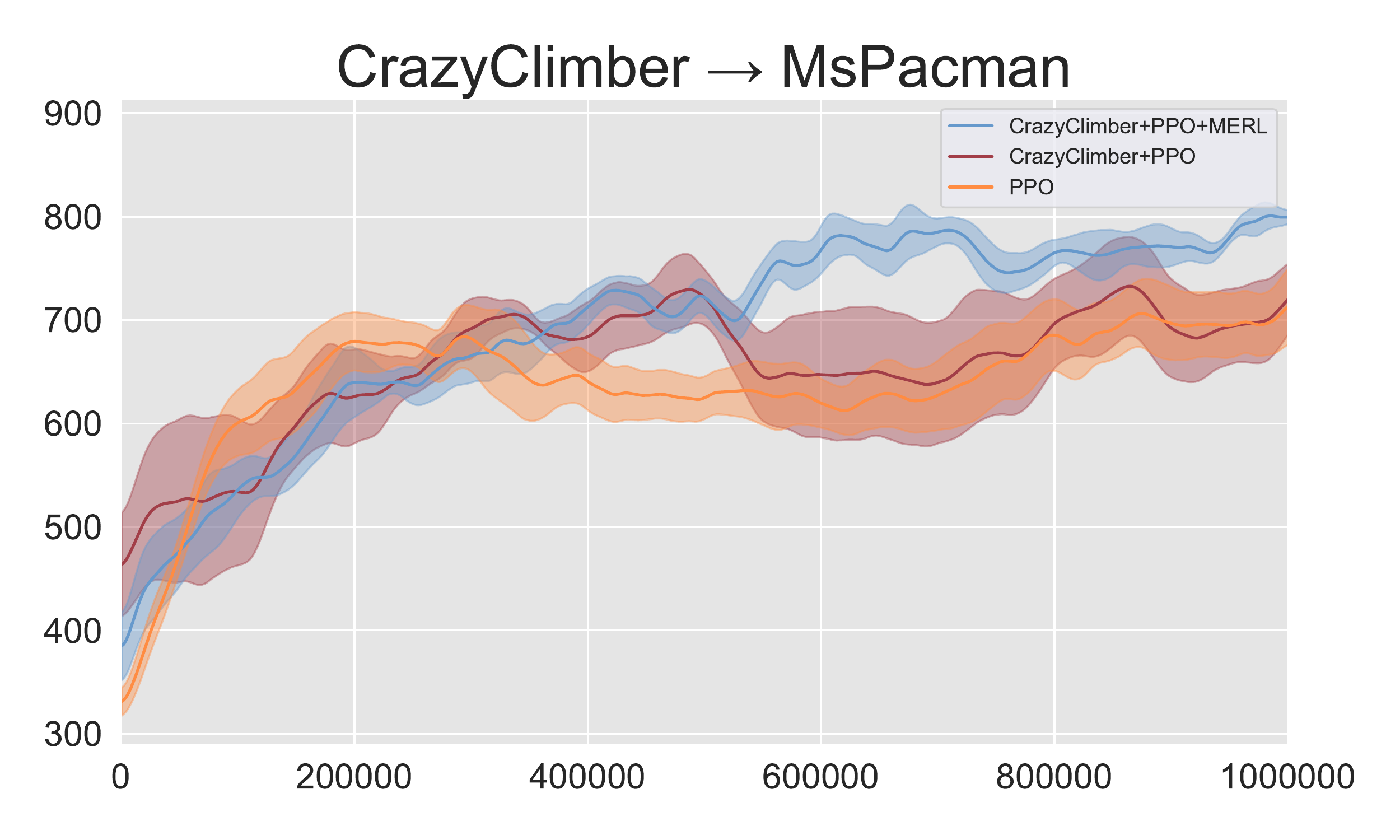}}}
    \qquad
    \subfloat{{\includegraphics[width=.33\linewidth]{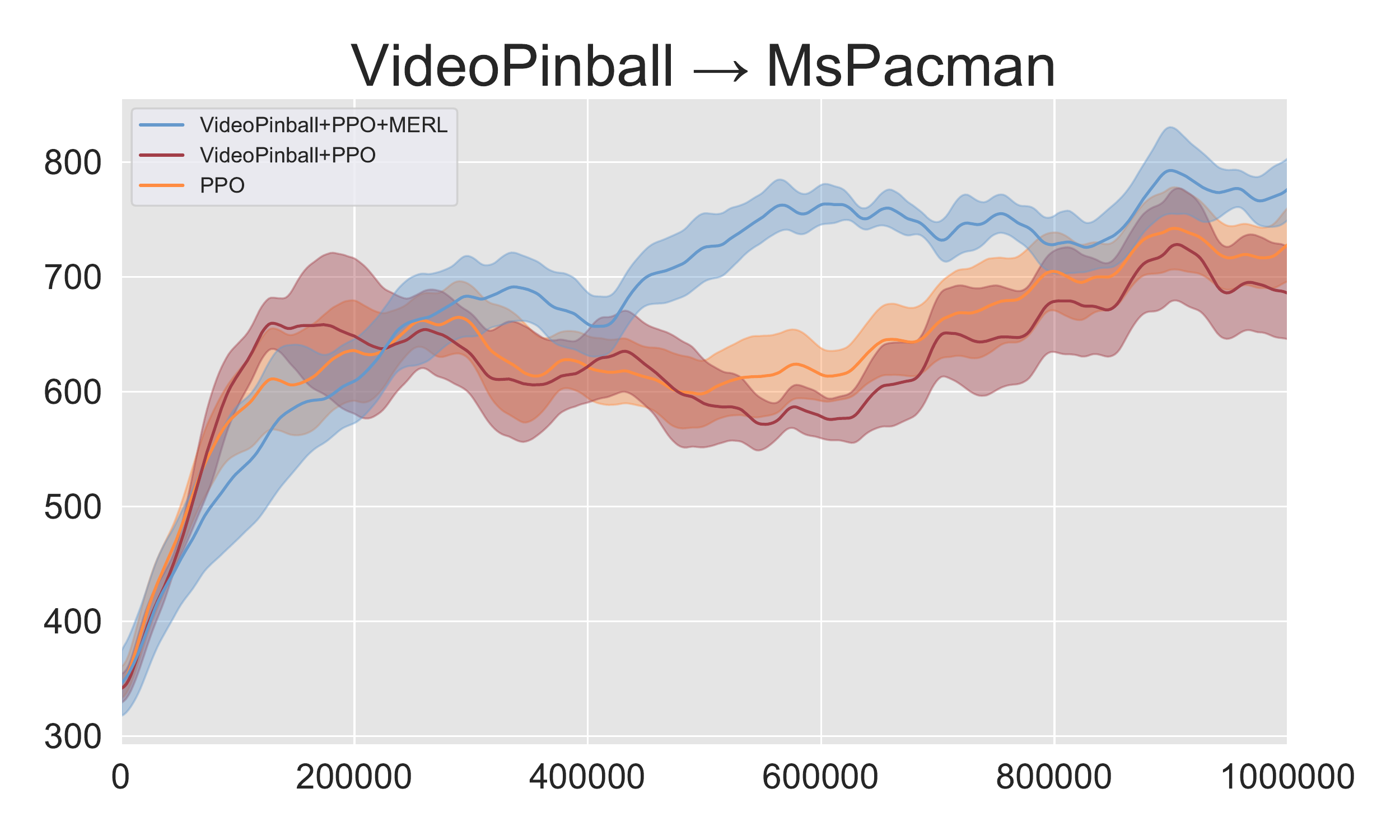}}{\includegraphics[width=.33\linewidth]{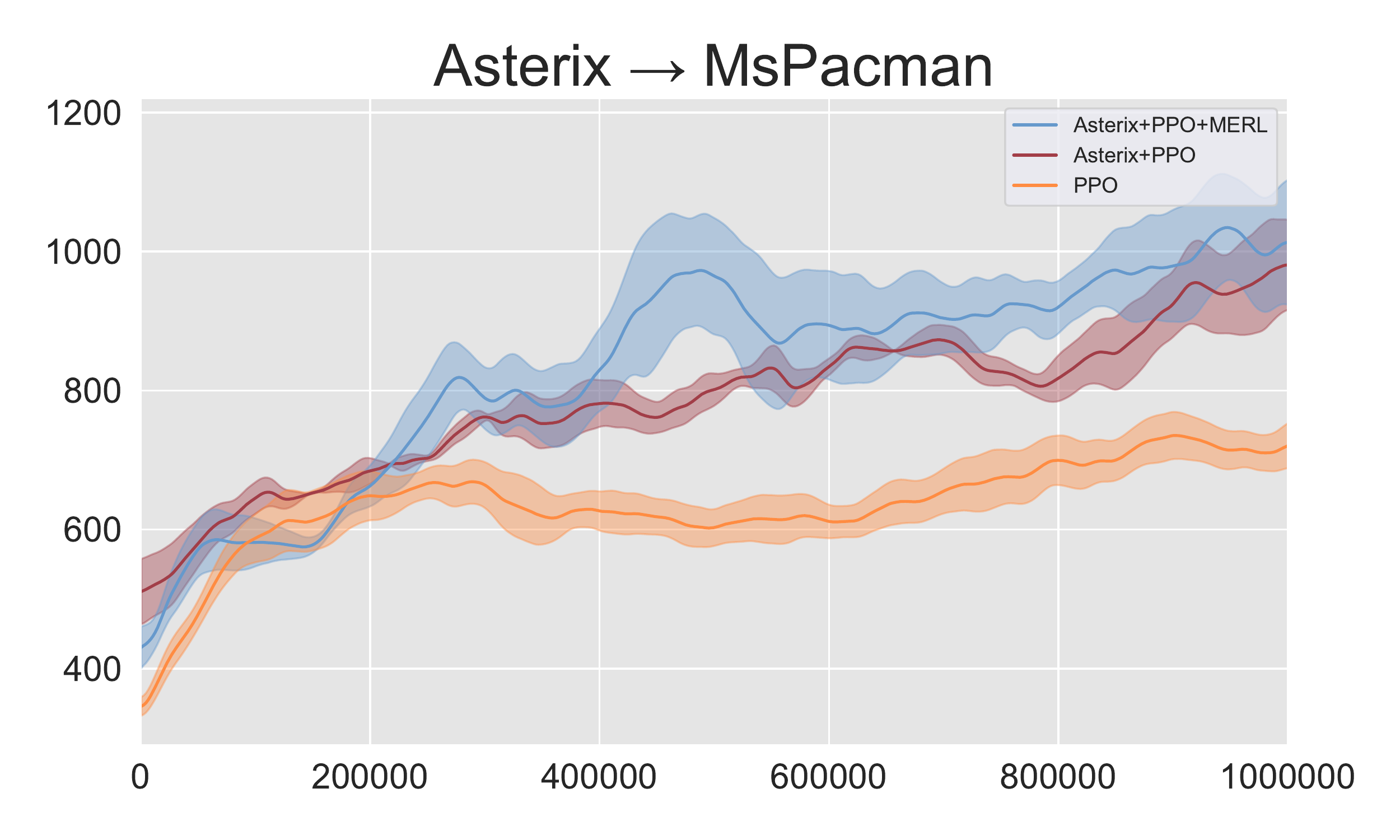}}}
    \caption{Transfer learning tasks from 5 Atari games to Ms. Pacman (2 $\times$ $10^6$ timesteps, 4 seeds). Performance on the second task. Orange is PPO solely trained on Ms. Pacman, red and blue are respectively PPO and our method transferring the learning. The line is the average performance, while the shaded area represents its standard deviation.}
    \label{fig:atari_ms}
\end{figure}

Fig.\@~\ref{fig:atari_ms} demonstrates that our method can better adapt to different tasks. This can suggest that MERL heads learn and help represent information that is more generally relevant for other tasks, such as self-performance assessment or accurate expectations. In addition to adding a regularization term to the objective function with problem knowledge signals, those auxiliary quantities make the neural network optimize for task-agnostic sub-objectives.

\subsection{Ablation Study}
\begin{figure}
    \centering
    \subfloat{{\includegraphics[width=.33\linewidth]{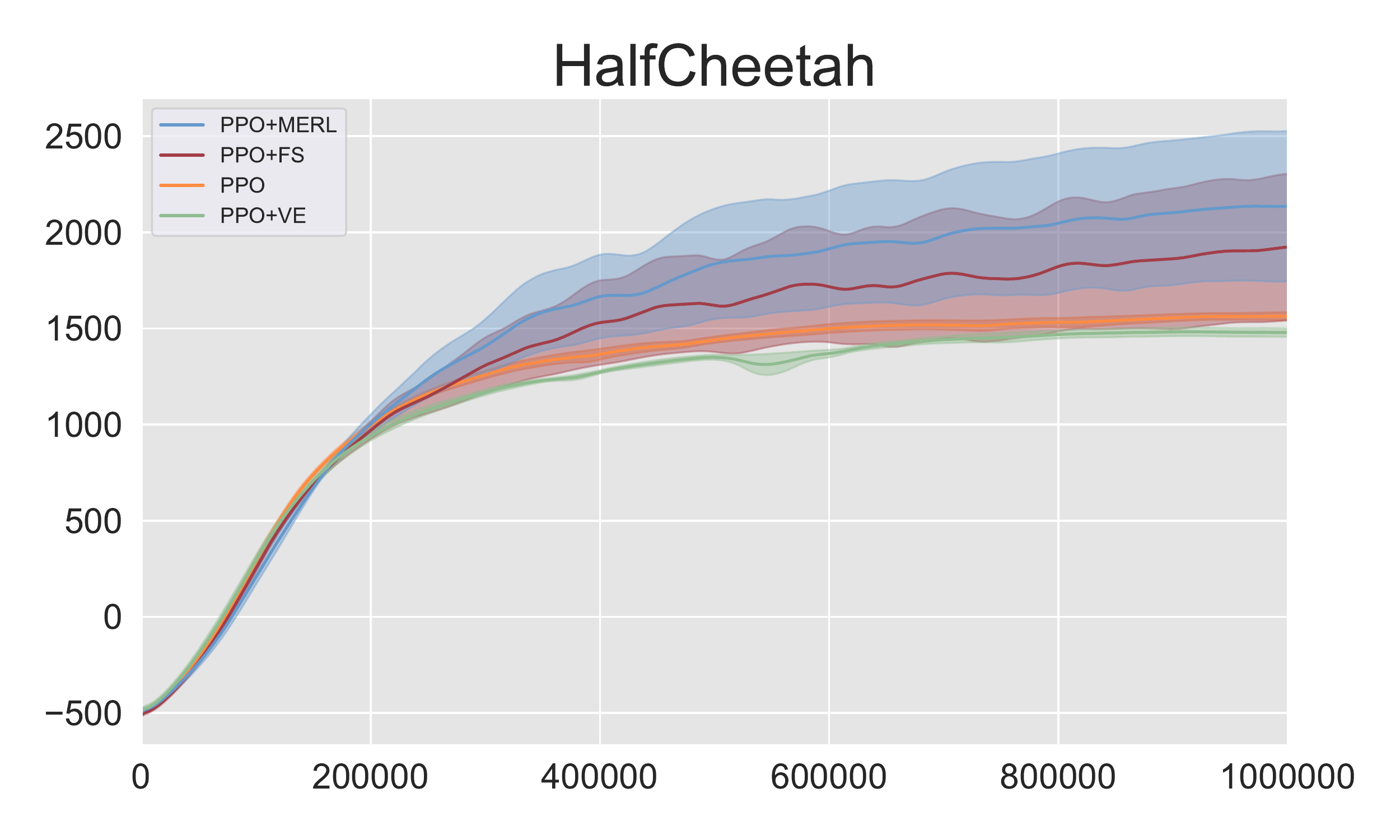}}{\includegraphics[width=.33\linewidth]{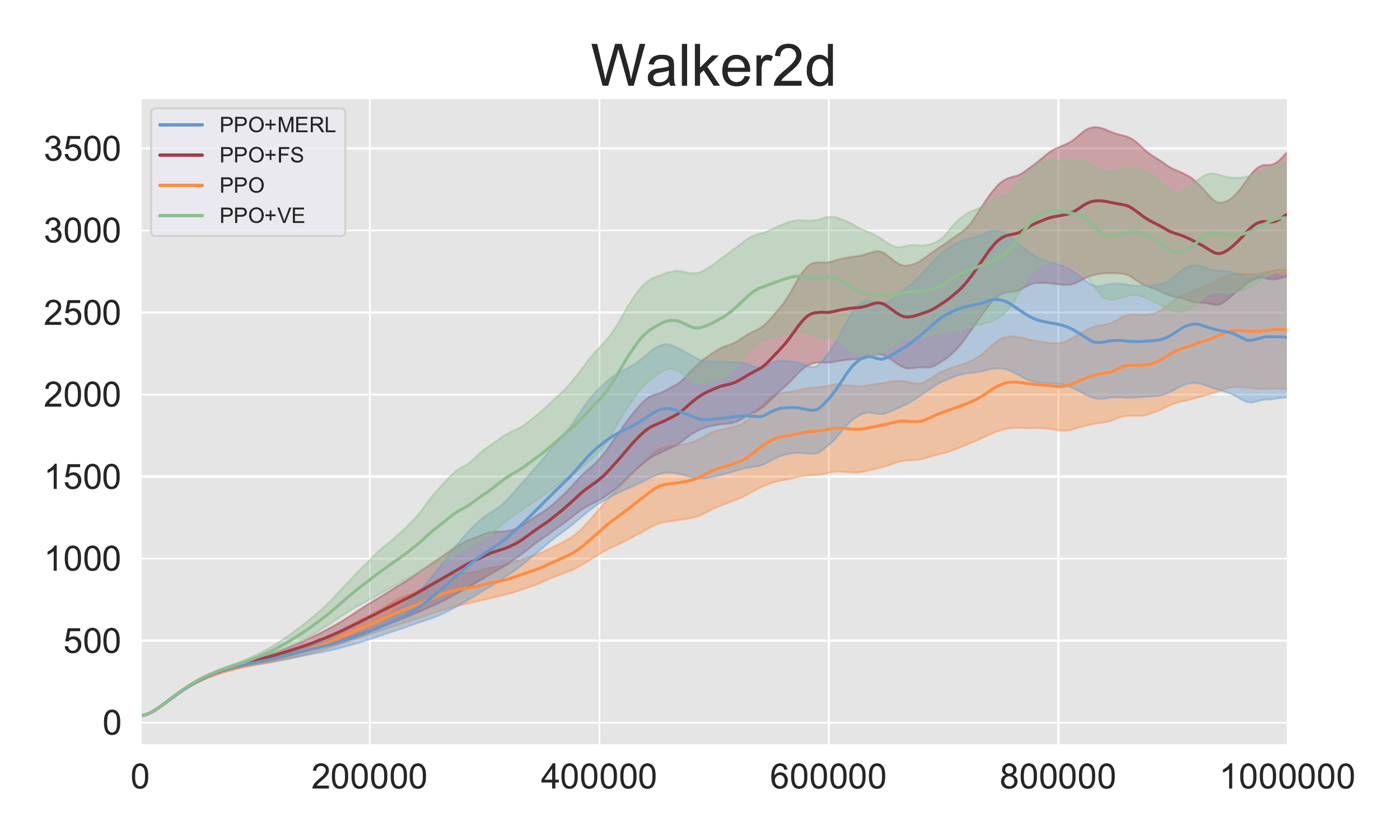}}{\includegraphics[width=.33\linewidth]{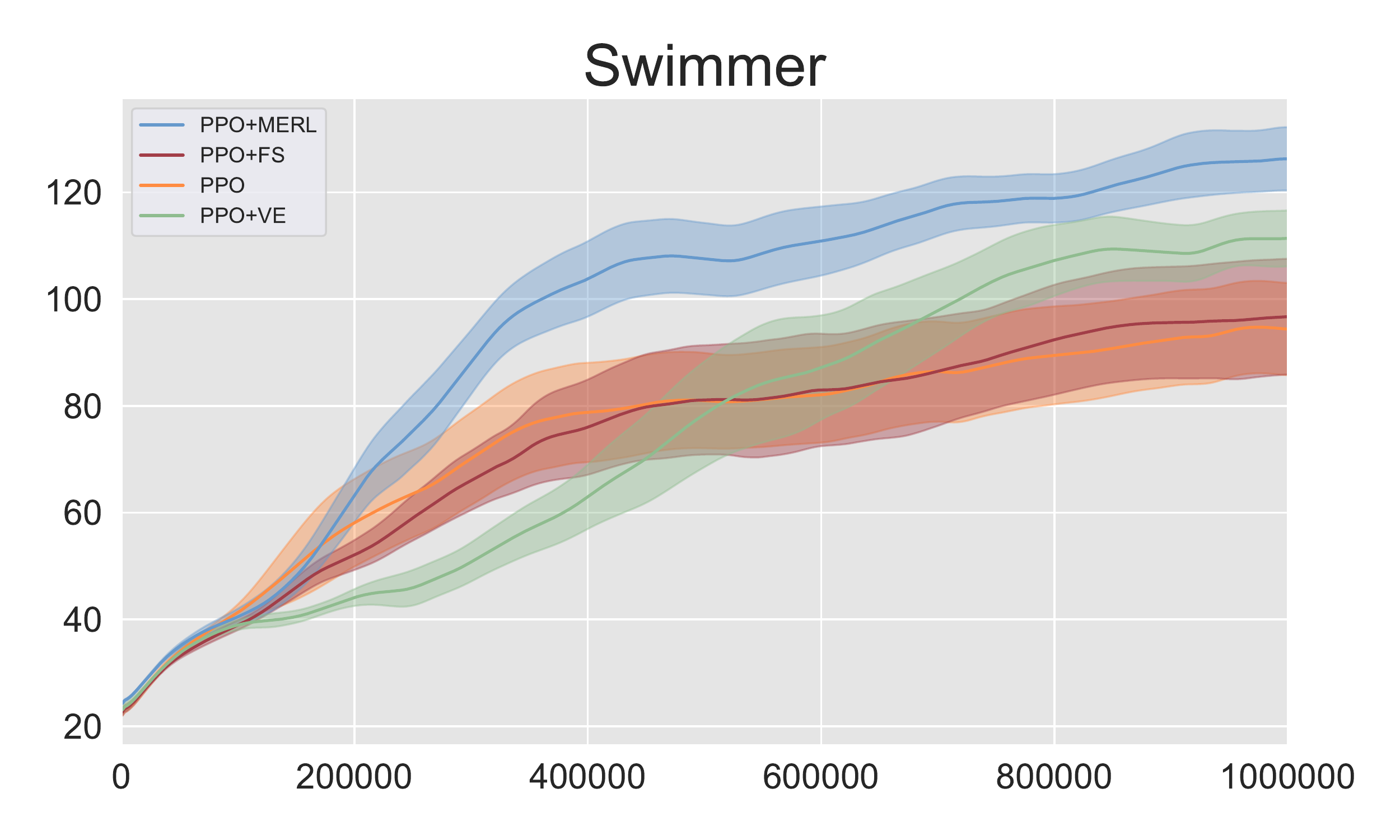}}}
    \caption{Ablation experiments with only one MERL head (FS or VE) ($10^6$ timesteps, 4 seeds). Blue is MERL with the two heads, red with the FS head, green with the VE head and orange with no MERL head. The line is the average performance, the shaded area represents its standard deviation.}
    \label{fig:ablation}
\end{figure}

We conduct an ablation study to evaluate the separate and combined contributions of the two heads. Fig.\@~\ref{fig:ablation} shows the comparative results in HalfCheetah, Walker2d, and Swimmer. Interestingly, with HalfCheetah, using only the $\mathrm{MERL}^{\mathrm{VE}}$ head degrades the performance, but when it is combined with the $\mathrm{MERL}^{\mathrm{FS}}$ head, it outperforms PPO+FS. Results of the complete ablation analysis demonstrate that each head is potentially valuable for enhancing learning and that their combination can produce remarkable results. In addition, it may be intuited that finding a variety of complementary MERL heads to cover the scope of the problem in a holistic perspective can significantly improve learning.

\subsection{Discussion}
The experiments suggest that MERL successfully optimizes the policy according to complementary quantities seeking for good performance and safe realization of tasks, i.e. it does not only maximize a reward but instead ensures the control problem is appropriately addressed. Moreover, we show that MERL is directly applicable to policy gradient methods while adding a negligible computation cost. Indeed, for the \textit{MuJoCo} and \textit{Atari} tasks, the computational cost overhead is respectively 5\% and 7\% with our training infrastructure. All of these factors result in a generally applicable algorithm that more robustly solves difficult problems in a variety of environments with continuous action spaces or by using only raw pixels for observations. Thanks to a consistent choice of complementary quantities injected in the optimization process, MERL is able to better align an agent's objectives with higher-level insights into how to solve a control problem. Besides, since many current methods involve that successful learning depends on the agent's chance to reach the goal by chance in the first place, correctly predicting MERL heads gives the agent an opportunity to learn from useful signals while improving in a given task.

\section{Conclusion}
In this paper, we introduced $\mathcal{V}^{ex}$, a new auxiliary task, to measure the discrepancy between the value function and the returns, which successfully assesses the agent's performance and helps learn more efficiently. We also proposed MERL, a generally applicable deep RL framework for learning problem-focused representations, which we demonstrated the effectiveness with a combination of two auxiliary tasks. We established that injecting problem knowledge signals directly in the policy gradient optimization allows for a better state representation that is generalizable to many tasks. $\mathcal{V}^{ex}$ provides a more problem-focused state representation to the agent, which is, therefore, not only reward-centric. MERL can be labeled as being a hybrid model-free and model-based framework, formed with lightweight embedded models of self-performance assessment and accurate expectations. MERL heads introduce a regularization term to the function approximation while addressing the problem of reward sparsity through auxiliary task learning. Those features nourish a framework technically applicable to any policy gradient algorithm or environment; it does not need to be redesigned for different problems and can be extended with other relevant problem-solving quantities, comparable to $\mathcal{V}^{ex}$.

Although the relevance and higher performance of MERL have only been shown empirically, we think it would be interesting to study the theoretical contribution of this framework from the perspective of an implicit regularization of the agent's representation on the environment. We also believe that the identification of additional MERL quantities (e.g., prediction of time until the end of a trajectory) and the effect of their combination is also a research topic that we find most relevant for future work.

%
%
\bibliographystyle{biblio}
\bibliography{biblio}

\end{document}